\documentclass{article}

\usepackage{appendix}
\usepackage{graphicx}
\usepackage{amssymb}
\usepackage{subcaption} 
\usepackage{float}  
\usepackage[skins,breakable]{tcolorbox}
\usepackage[preprint]{corl_2024} 

\title{Trajectory Adaptation Using Large Language Models}

\author{
 Anurag Maurya\\
   Robert Bosch Centre for Cyber-Physical Systems\\ 
 Indian Institute of Science, Bengaluru, 
  India\\
  \texttt{anuragm1@iisc.ac.in} \\
\And
  Tashmoy Ghosh\\
  Robert Bosch Centre for Cyber-Physical Systems\\ 
 Indian Institute of Science, Bengaluru,
  India\\
  \texttt{tashmoyg@iisc.ac.in} \\
\And
 Ravi Prakash\\
  Robert Bosch Centre for Cyber-Physical Systems\\ 
 Indian Institute of Science, Bengaluru,
  India\\
  \texttt{ravipr@iisc.ac.in}} 
\begin{document}
\maketitle


\begin{abstract}
     Adapting robot trajectories based on human instructions as per new situations is essential for achieving more intuitive and scalable human-robot interactions. This work proposes a flexible language-based framework to adapt generic robotic trajectories produced by off-the-shelf motion planners like RRT, A-star, etc, or learned from human demonstrations. We utilize pre-trained LLMs to adapt trajectory waypoints by generating code as a policy for dense robot manipulation, enabling more complex and flexible instructions than current methods. This approach allows us to incorporate a broader range of commands, including numerical inputs. Compared to state-of-the-art feature-based sequence-to-sequence models which require training \cite{C4} \cite{C8}, our method does not require task-specific training and offers greater interpretability and more effective feedback mechanisms. We validate our approach through simulation experiments on the robotic manipulator, aerial vehicle, and ground robot in the Pybullet and Gazebo simulation environments, demonstrating that LLMs can successfully adapt trajectories to complex human instructions.
\end{abstract}



\section{Introduction}
%


Trajectory adaptation in robotics presents several challenges, particularly in planning-based methods \cite{C19} that often require replanning for every new situation. This can become computationally expensive and prone to errors in real-time applications. In contrast, learning-based models offer faster adaptation by leveraging task parameters, such as those used in TP-GMM or dynamical systems like Elastic Dynamical System Motion Policies \cite{tpgmm} \cite{dytg}, to adjust the learned model to new environments. However, these approaches require precise mathematical modeling and the explicit formulation of task frames. In our work, we explore a natural language-based trajectory adaptation method that relies on high-level plans and policies encoded as code. This approach bypasses the need for complex mathematical formulations, making it more intuitive, fast, and scalable.

Large Language Models (LLMs) can assist by enabling natural language-based adaptations to trajectories, allowing non-expert users to suggest changes in intuitive ways. Translating human commands to low-level control \citep{C13}, particularly towards adjusting a trajectory based on specific preferences like "Go higher" or "Move slower" has been challenging due to the gap between language interpretation and task adaptation.
LLMs have shown great capability in understanding commands and generating rough plans for executing instructions. They have also demonstrated consistency in generating code \cite{C15}. Code has become a widely used medium to connect high-level plans with low-level trajectory adjustments.

We build on the concept of using code as a bridge, extending it to adapt existing trajectories in alignment with given commands. Trajectories can be understood as a series of waypoints. When adapting a trajectory, the goal is to adjust certain key waypoints while maintaining the overall characteristics, such as the trajectory's shape and its smoothness. Our goal is to interpret the user's intent as a series of required adaptations and then utilize code as the means to apply those changes. For example, "Walk at a distance of at least 10 from the fan" can be interpreted as a sequence of identifying points closer to the fan and then increasing their distance from the fan. Additionally, the user provides feedback to clarify and fully articulate the requirements, helping to eliminate any assumptions or potential errors.


\section{Related Works}
\label{sec:citations}

 \textbf{Natural Language and Robotics:}
Providing natural language models for robots offers a simple and user-friendly interface for addressing these issues through decision-making and human interaction. Traditionally, it has been difficult to represent human-robot interactions using language because it needs the user to follow a strict set of instructions \cite{C1} or involves sophisticated mathematical methods to track various probability distributions over actions and target objects \cite{C2}, \cite{C3}. There has been an increase in recent works that explore Large Language Models (LLM) for complex mapping between language and actions over embodied agents as discussed in \cite{C10},\cite{C11}. It covers the essential challenges of applying LLMs to robotics, such as the need for real-time responses, handling multi-modal inputs (e.g., language, vision, and proprioception), safety concerns, and the difficulty of grounding language in physical environments. This challenge provides a base for our research in developing a robust natural language-oriented robot task manipulation pipeline. We focus on engineering three essential areas, the prompt designing, to provide essential numerical understanding rules to LLMs, utilizing code as a Adaptation policy, and an iterative closed loop feedback system with a corresponding high-level plan that can be inferred by a user for required corrections.

\textbf{Trajectory Adaptation with Natural Language:}
Several studies have explored multi-modal representations that combine natural language with environmental state observations for trajectory Adaptation. Notably, a transformer-based model called LaTTe \cite{C4},\cite{C5}, particularly focused on solving trajectory-based tasks, where, given an initial trajectory, the robot adjusts it within a continuous action space to follow a specified sequence or path based on language commands. However, this approach requires training and is biased towards a limited set of simple instruction-based adaptations. We propose an approach with no training requirement and capable of handling complex multi-step instructions with numerical-oriented Adaptation.
 
\textbf{Large Language Models for Robotics: }
In the area of large pre-trained language models (LLMs) applied to robotics, models like GPT are being directly utilized to generate robotic trajectories without the need for task-specific fine-tuning, often in a zero- or few-shot setting \cite{C6} \cite{C17} where a language model is prompted with natural language descriptions of a task and produces a sequence of movements or positions in a higher-level action space by generating code to be used as a policy \cite{C7} with pre-trained LLM.Unlike traditional end-to-end models that directly output trajectories or commands, code generated by language models includes high-level logic, enabling robots to handle complex tasks with reasoning and error recovery. While \cite{C7} explores if waypoints can be generated to achieve the execution of a task, our work explores how we can utilize a similar approach to adapt an existing trajectory as per user's preferences \cite{C12} utilizes a closed-loop strategy with source feedback such as success detection, scene description, and human interaction \cite{C18}. Various methods develop an online explainable trajectory correction for robots based on natural language inputs \cite{C8} \cite{C9} by interpreting human commands that describe desired adaptations to a robot's trajectory and generate explainable adjustments based on interpretable features such as speed, direction, and proximity to objects. The primary focus of our research is to adapt human-demonstrated dense trajectories based on numeric and object-specific instructions. This process is guided by an iterative feedback system, where the user refines the structured high-level plan to make necessary modifications.


\section{Method}
\label{sec:Method}
\subsection{Problem Formulation}
We describe the task of adapting the trajectory as a mapping from the initial set of waypoints, influenced by environmental context and language-based instructions. Let $\tau : \mathbb{R} \to \mathbb{R}^4$ represent the original trajectory, composed of $N$ waypoints and corresponding velocities, $\tau = \{(x_1, y_1, z_1, v_1), \dots, (x_N, y_N, z_N, v_N)\}$, where $x_i$, $y_i$, $z_i$, and $v_i$ denote the position coordinates and velocity at each time step $i$. This initial trajectory is assumed to have been generated using standard algorithms or previously demonstrated by a human expert.

Let $L_{instruct}$ be the command reflecting a user's intention to adjust the trajectory based on their preferences, for example: "Slow down when near the person." These commands may express multiple preferences but are assumed to be sufficiently clear in conveying the user’s intent. Additionally, $O$ is defined as an observation set representing $m$ objects in the environment, each with a position $P(O_i) \in \mathbb{R}^3$ and label $Label(O_i)$, and $E_d$ as a natural language description of the environmental semantics and rules. Our goal is to learn a function $g$ that maps the original trajectory, observations, and user instructions to produce an adapted trajectory $\tau_{mod}$ that aligns with the intent expressed in $L_{instruct}$.
\[
\tau_{mod} = g_{LLM}(\tau, L_{instruct}, O)
\]

\subsection{Methodology}
 We leverage the capabilities of LLMs to approach the problem outlined in Section 3.1. A parallel process of simultaneously generating a High-Level Plan (HLP) and a Python Code is described as follows:\newline
\subsubsection{HLP Generation}
First, the LLM interprets the user’s natural language instructions and the environmental description to generate a step-by-step high-level plan. This plan outlines how the trajectory should be adapted, considering the user's preferences, and the contextual information provided. Examples of HLP generated by our approach can be viewed in the appendix \ref{HLP generated} \newline
\subsubsection{Code Generation}
Next, consistent with this high-level plan, the LLM generates executable code. This code performs the necessary operations on the initial trajectory which aligns with the user's intent. By automating both the planning and coding phases, the LLM acts as a bridge between high-level user instructions and low-level code execution. 


After the high-level plan and code are generated, the plan is sent to the user for validation. If the user confirms its correctness, the code is executed and the adapted trajectory is forwarded to the execution module. If the user identifies any issues or omissions in the high-level strategy, they provide feedback. This feedback, along with the original instruction, is fed back into the framework to produce updated results that align with the user's input. We have used two in-context examples of high-level plans in order to make them suitable for human debugging, listed in appendix \ref{In-context examples}

\begin{figure}[h]
    \centering
    \includegraphics[width=\linewidth]{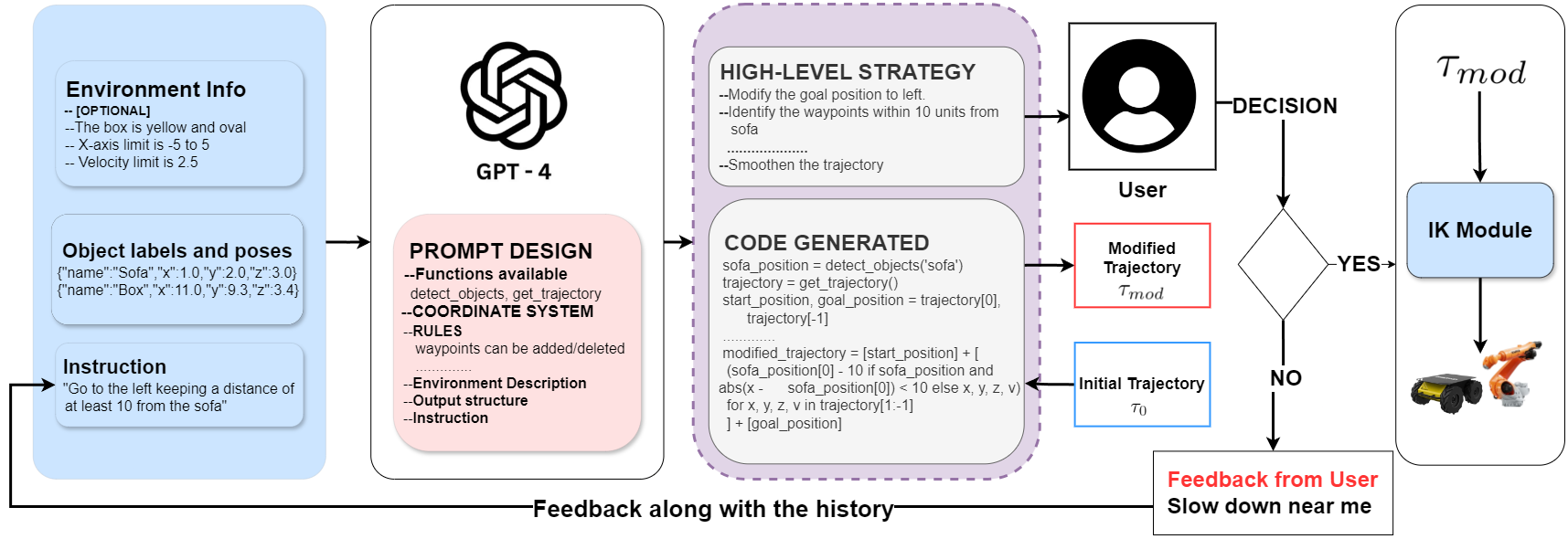} 
    \caption{An overview of the pipeline: A prompt is created from environment description, instruction, and task agnostic guidelines; LLM generates high-level plan along with Python code. The user reviews the plan; if correct, the code runs to adapt the trajectory. If a feedback is given, it's sent back to the LLM with the original instruction.  }
    \label{fig:methodology}
\end{figure}

\newpage
\section{Experiments}
\label{sec:Experiments}
Our experiments aim to address the following key questions:
\begin{enumerate}
    \item Can LLMs be applied to generate code for transforming trajectories which includes waypoints as well as velocity components?
    \item  With just one or two examples of high-level strategies, can LLMs devise plans for more intricate and diverse commands to adapt the trajectory?? 
    \item How can our approach be used for free-form (open-ended, natural language commands that are not restricted by predefined syntax or rigid formats)manipulation of the waypoints and give an advantage over the State-of-the-art methods? 
\end{enumerate}

\subsection{Dataset} \label{Dataset}
Each data sample includes an initial trajectory $\tau_0$, environmental observations comprising object labels $Label(O_i)$, object poses $P(O_i)$, an optional linguistic description of the environment $E_d$, and a natural language instruction $L_{instruct}$. The environment description provides valuable information that can be used to interpret the commands effectively. 

We utilize a subset of the already available dataset from the paper LaTTe (Language Trajectory Transformer) \cite{C4}. This dataset primarily captures three major types of instructions: 
\begin{enumerate}
\item Altering waypoints in the Cartesian trajectory space, for example, "Go to the left", or "Move to the front", 
\item  Adjusting speed, for example, "Increasing the speed near the person", "Go slower", and 
\item Positional shifts with reference to objects, for example, "Drive closer to the sofa", "Stay further away from box". 
\end{enumerate}
The LaTTe dataset focuses on general changes rather than exact numerical ones. Also, the commands are simple and don’t include multiple instructions at once. This subset excludes commands like "Upper part" and "Top"  because they are not informative enough for a pre-trained model.

We create a new set that addresses these problems and includes complex language commands along with numerical changes. for example, "Go left by 20", and "Keep at least 10 distance from the box". We extend this to include multiple instructions in a single command, for example, "Go left by 20 keeping a distance of at least 10 from the box". A sample list of commands can be found in the appendix. \ref{appendixA}.

\subsection{Prompt Design}
The core idea of our approach is to explore how LLMs can define formatting functions as code and understand how to adapt trajectories in accordance with instructions. We define the output generated by LLM to be in two processes: First, a high-level plan that can be reviewed by the User, and Second, a code that adapts the trajectory. The key considerations in designing the prompt are as follows:

\subsubsection{ How do LLMs know the environment configuration?} Natural commands, for example, "Go to the left/right" do not provide enough information about the direction in which change should be done. We provide a notion of COORDINATE SYSTEM along with optional environment descriptors in the prompt which can be set by the user as per the required setup. Additionally, We provide the LLM with the definition of two functions, detect\_objects(obj\_name) which returns the position of an object in the environment, and get\_trajectory() which returns the trajectory as a set of waypoints along with velocity.

\subsubsection{Defining desired characteristics in a trajectory using rules} The LLM has to determine if the goal position or starting position will remain the same based on the instruction. Furthermore, The shift in the waypoints should be gradual to ensure smooth transitions. Lastly, the LLM is given the choice to add or remove any number of waypoints as required. Through experiments, we designed a prompt that uses two in-context examples  \ref{In-context examples} of high-level plan to ground the response to our intent. \textbf{The full prompt can be found in appendix \ref{appendixA}.}
 
\subsection{Simulation experiments}
\begin{figure}[h]
    \centering
    \begin{subfigure}[b]{0.3\textwidth} 
        \centering
        \includegraphics[width=\textwidth]{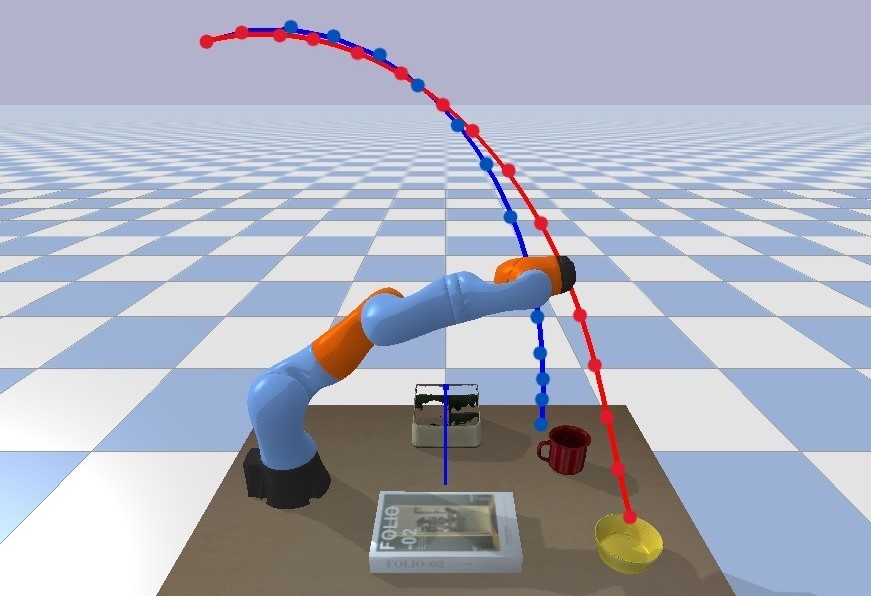}
        \caption{Shift the trajectory to reach yellow bowl}
    \end{subfigure}
    \begin{subfigure}[b]{0.3\textwidth}
        \centering
        \includegraphics[width=\textwidth]{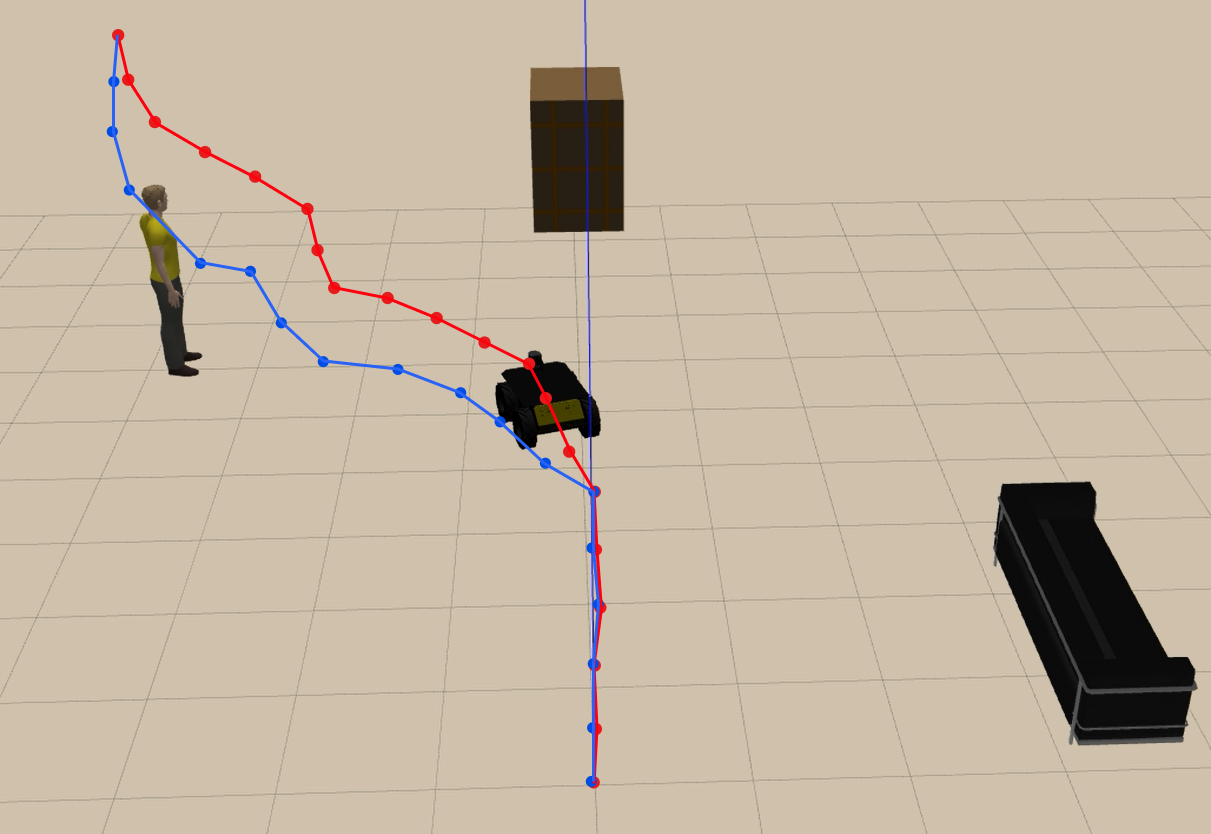}
        \caption{Drive at a greater distance from person}
    \end{subfigure}
    \begin{subfigure}[b]{0.3\textwidth}
        \centering
        \includegraphics[width=\textwidth]{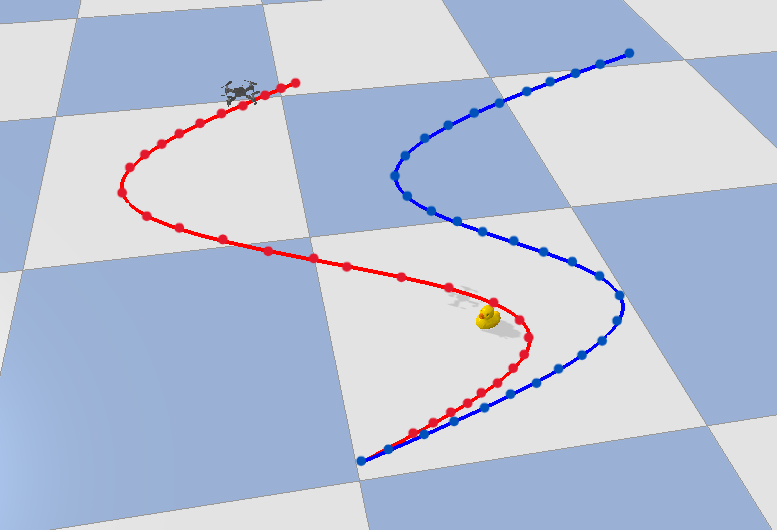}
        \caption{Go towards left passing closer to the duck}
    \end{subfigure}
    \caption{Simulation experiments: Testing across varied robot dynamics, a) Robotic arm, b) Ground robot, c) Drone. The initial trajectory is represented in Blue and the adapted trajectory is shown in Red }
    \label{fig:experimets}
\end{figure}

To create an assessment of our approach's performance and utility to varying robot dynamics and surroundings, we test our approach across varying simulators, robots, and multiple tasks. We tested our framework on a ground robot in the Gazebo \cite{gazebo}, the crazy fly drone \cite{panerati2021learning}, and the Kuka robotic arm \cite{kuka} in the Pybullet, as shown in Figure \ref{fig:experimets}. The experiments demonstrate uniform object-oriented trajectory shifting as shown in Fig 2 (a), generic shift concerning simple words describing things to varying degrees in (b), and directional shifting in (c) This provides us an insight into how different platforms and robot dynamics, low-level IK modules can be used together with our framework. All experiments use the GPT-40 model with a temperature of 0.1

\subsection{Results and Discussion}
\begin{figure}[h]
    \centering
    \includegraphics[width=\textwidth]{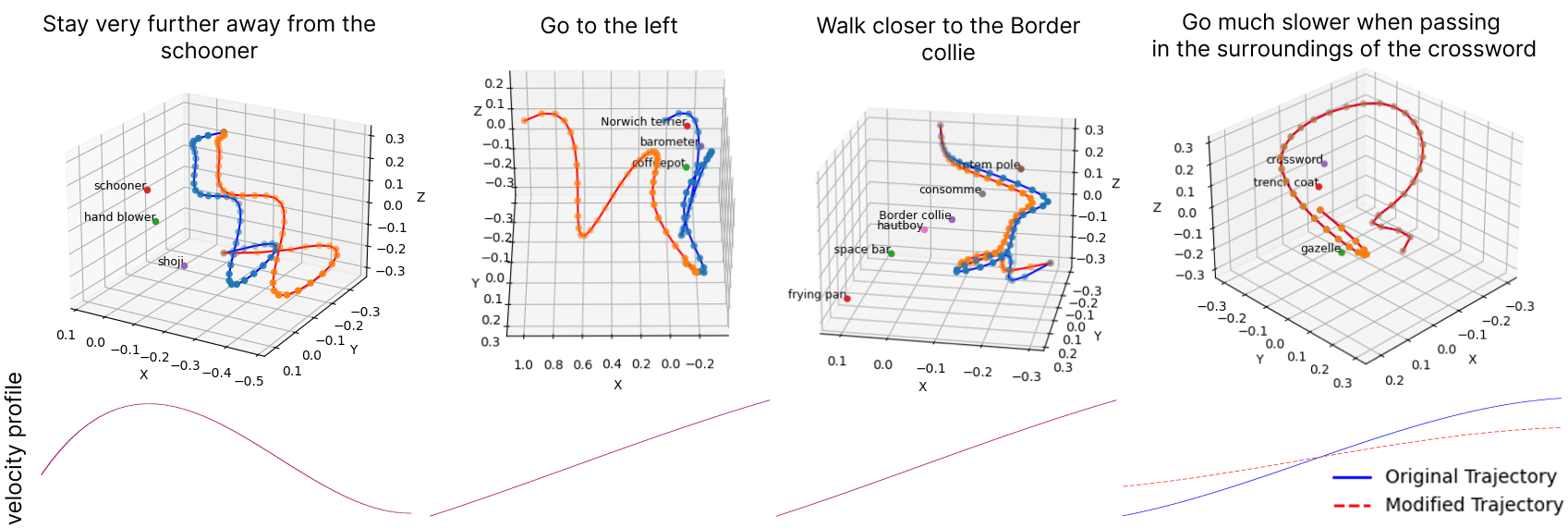} 
    \caption{Results over LaTTe Dataset}
    \label{fig:results_latte}
\end{figure}

We examine our approach on a subset of the LaTTe dataset, as described in section \ref{Dataset}. The results, shown in figure \ref{fig:results_latte}, validate our method as an effective LLM-based alternative to existing approaches. As previously mentioned in section \ref{Dataset}, we excluded commands that lacked sufficient detail for meaningful analysis.  

\begin{figure}[h]
    \centering
    \includegraphics[width=\textwidth]{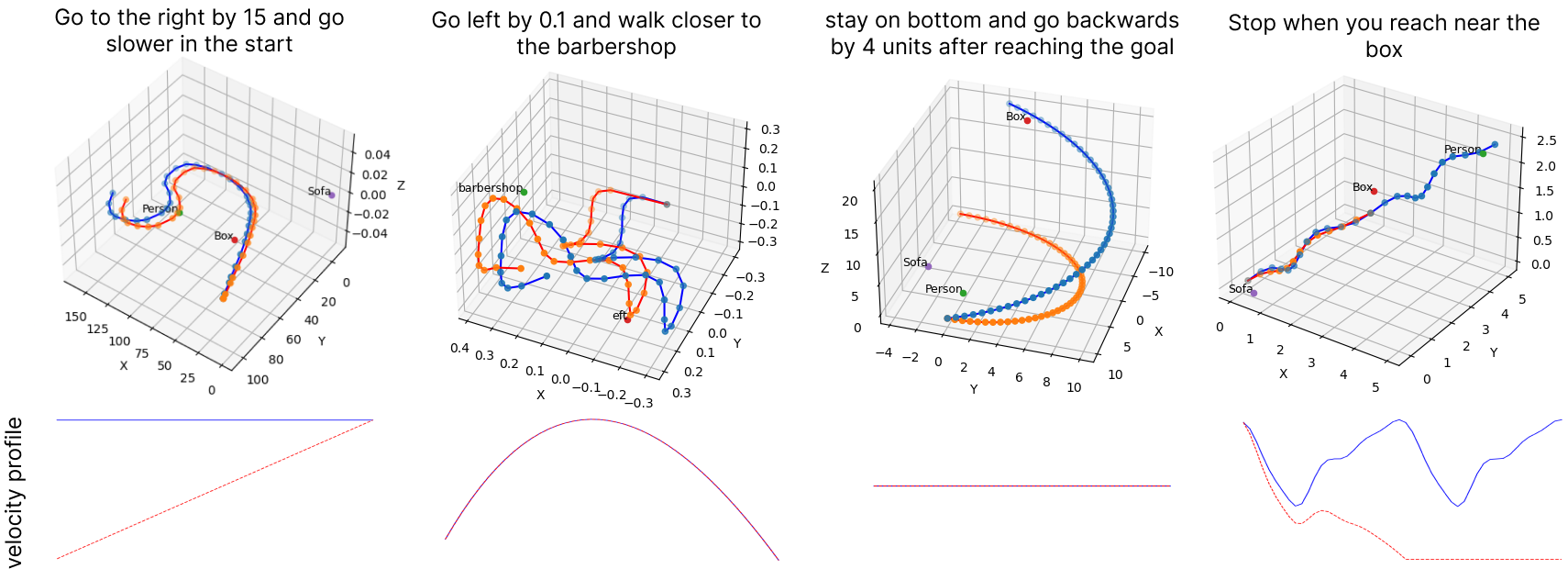} 
    \caption{Results over extended dataset having complex instructions along with numerical changes}
    \label{fig:results_extended}
\end{figure}

\textbf{Can LLMs be applied to generate code for transforming trajectories which includes waypoints as well as velocity components?} The adapted trajectory generated by the code aligns with the intended meaning of the instruction. The trajectory generated by the code has smooth adjustments that maintain a shape similar to the original path as shown in Figure \ref{fig:results_latte} and Figure \ref{fig:results_extended}

\textbf{With just one or two examples of high-level strategies, can LLMs devise plans for more
intricate and diverse commands to adapt the trajectory?}
Moving beyond simple commands, we also tested our approach on a custom dataset that includes more complex instructions with both precise numerical values and compound instructions. Figure \ref{fig:results_extended} illustrates the results obtained on the extended dataset. Appendix \ref{HLP generated} provides examples of high-level strategies generated for a few complex instructions. We found that LLMs are capable of understanding new modifications and generating the appropriate code to implement them.

\begin{figure}[h]
    \centering
    \includegraphics[width=\textwidth]{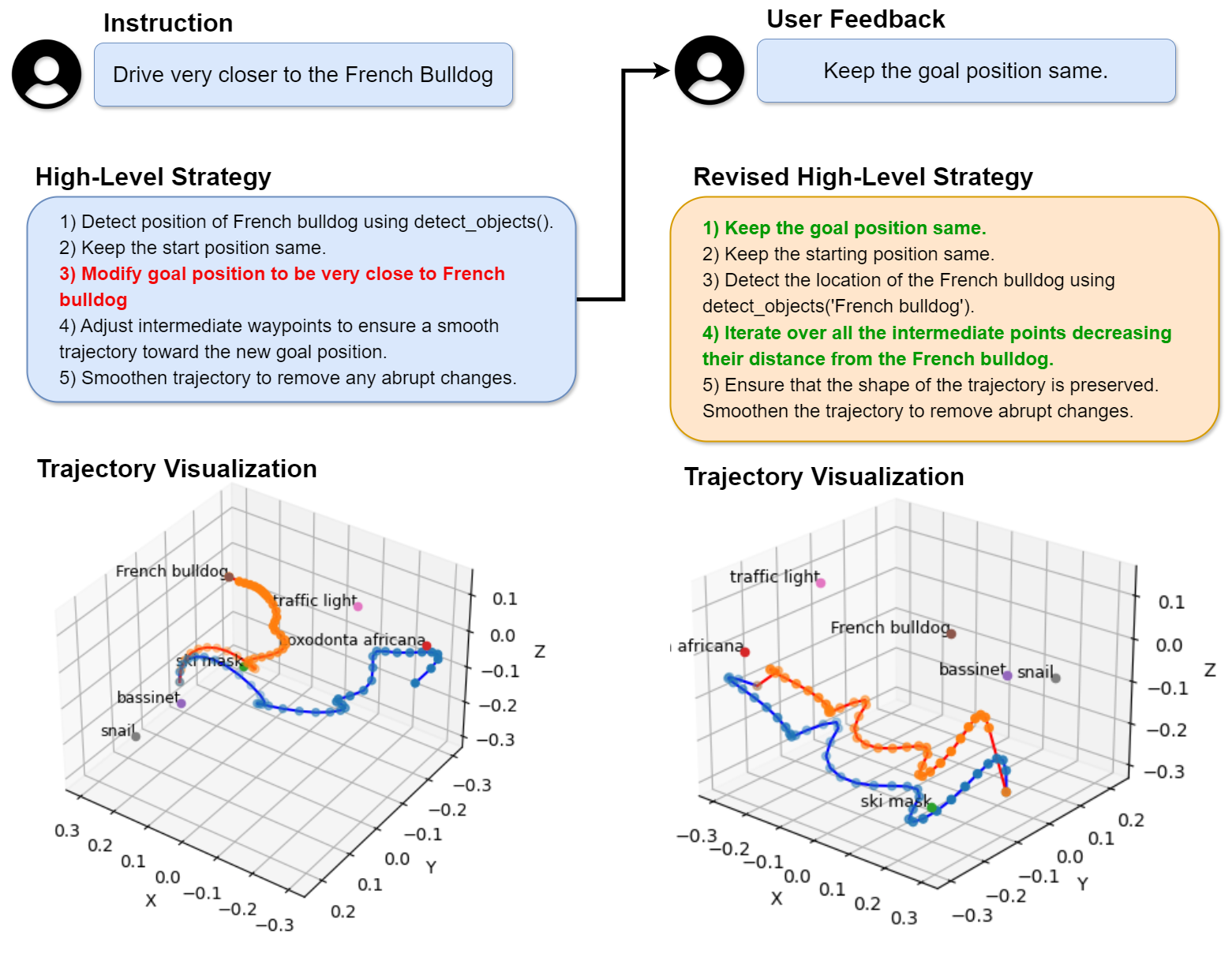} 
    \caption{Providing additional information to LLMs, which may be missing from the high-level strategy, helps correct inconsistencies and ensures more accurate outputs}
    \label{fig:feedback_diag}
\end{figure}

\textbf{How can our approach be used for free-form (open-ended, natural language commands that
are not restricted by predefined syntax or rigid formats)manipulation of the waypoints and
give an advantage over the State-of-the-art methods?}
In figure \ref{fig:feedback_diag}, we further demonstrate how the system can handle misinterpretations by integrating user feedback, allowing for clarifications regarding the intent to correct the trajectory. This extended assessment showcases the versatility of our framework in handling both simple and complex commands while providing an effective way for error diagnostics.  We found that the generated code consistently followed the user’s instructions for specific parts of the trajectories as well. For instance, with the instruction "Stop when you reach near the box,", Figure \ref{fig:results_extended} the method identifies the closest waypoint to the box and halts at that point.



\section{Limitations}
Though our work enables conversations with an LLM to produce instruction-compliant trajectories,  the effectiveness of the adaptations depends on how descriptive and complete the instructions are. Terms like 'very', 'very much', and 'relatively slower' are very subjective and can vary from person to person. The User needs to be very descriptive in how the commands are conveyed and if the information supplied is enough. Additionally, our method lacks mathematical guarantees, which could lead to semantically accurate but irregular or non-smooth transitions in the trajectories. Moreover, due to open-vocab instructions, it might not be possible to tailor a single constraint satisfaction module for all possible scenarios. 

\section{Conclusion}
This work presents a method for efficient language-based human-robot interaction for trajectory Adaptation. We leverage the pre-trained LLM GPT-4o for language-conditioned trajectory manipulation. Our method incorporates providing LLM with relevant information required like object positions, environment description, and natural language instruction which is further employed for generating a consecutive high-level plan and code for numerical adaptation.

The approach can be applied to diverse platforms such as arm manipulators, aerial vehicles, and ground robots. The simulation experiments demonstrate the capabilities of our pipeline for free-form task-instructed trajectory reshaping without any fine-tuning. Our method can incorporate more complex instructions that can include numerical inputs. The language used can be complex. A closed-loop system with human feedback ensures quality control and corrects any misinterpretations or errors by the LLM.

 Our future directions will extend to experimenting with more complex instructions and sequential long-horizon tasks exploring the implications of LLMs. Furthermore, it might be desirable to have dataset bias in some use cases. We believe that our framework can serve as an important direction for paradigms in human-robot collaboration that employ large language models. We intend to conduct a user study to evaluate the method's effectiveness and ease of use.
 
\clearpage
\acknowledgments{If a paper is accepted, the final camera-ready version will (and probably should) include acknowledgments. All acknowledgments go at the end of the paper, including thanks to reviewers who gave useful comments, to colleagues who contributed to the ideas, and to funding agencies and corporate sponsors who provided financial support.}


\bibliography{example}  

\begin{thebibliography}{22}
\providecommand{\natexlab}[1]{#1}
\providecommand{\url}[1]{\texttt{#1}}
\expandafter\ifx\csname urlstyle\endcsname\relax
  \providecommand{\doi}[1]{doi: #1}\else
  \providecommand{\doi}{doi: \begingroup \urlstyle{rm}\Url}\fi

\bibitem[Bucker et~al.(2023)Bucker, Arthur, and et~al.]{C4}
Bucker, Arthur, and et~al.
\newblock Latte: Language trajectory transformer.
\newblock \emph{IEEE International Conference on Robotics and Automation (ICRA)}, IEEE 2023\penalty0 (4), 2023.

\bibitem[Yow and et~al.(2024)]{C8}
J.~Yow and et~al.
\newblock Extract--explainable trajectory corrections from language inputs using textual description of features.
\newblock \emph{arXiv preprint}, arXiv:2401.03701\penalty0 (8), 2024.

\bibitem[LaValle(2006)]{C19}
S.~LaValle.
\newblock Planning algorithms.
\newblock \emph{Cambridge university press,}, \penalty0 (17), 2006.

\bibitem[Calinon(2015)]{tpgmm}
S.~Calinon.
\newblock Robot learning with task-parameterized generative models.
\newblock In \emph{Proc. Intl Symp. on Robotics Research ({ISRR})}, 2015.

\bibitem[Li and Figueroa(2023)]{dytg}
T.~Li and N.~Figueroa.
\newblock Task generalization with stability guarantees via elastic dynamical system motion policies.
\newblock In \emph{7th Annual Conference on Robot Learning}, 2023.
\newblock URL \url{https://openreview.net/forum?id=8scj3Y0RLq}.

\bibitem[Hunt et~al.(2024)Hunt, Ramchurn, and Soorati]{C13}
W.~Hunt, S.~D. Ramchurn, and M.~D. Soorati.
\newblock A survey of language-based communication in robotics.
\newblock \emph{arXiv preprint}, arXiv:2406.04086v1\penalty0 (13), 2024.

\bibitem[JIANG et~al.(2024)JIANG, WANG, and et~al.]{C15}
J.~JIANG, F.~WANG, and et~al.
\newblock A survey on large language models for code generation.
\newblock \emph{arXiv preprint}, arXiv:2406.00515\penalty0 (15), 2024.

\bibitem[Tellex et~al.(2020)Tellex, Gopalan, Kress-Gazit, and C.]{C1}
S.~Tellex, N.~Gopalan, H.~Kress-Gazit, and M.~C.
\newblock Robots that use language.
\newblock \emph{Annual Review of Control, Robotics, and AutonomousSystems}, 3\penalty0 (1):\penalty0 25–55, 2020.

\bibitem[Arkin et~al.(2020)Arkin, Park, Roy, Walter, Roy, Howard, and R.]{C2}
J.~Arkin, D.~Park, S.~Roy, M.~Walter, N.~Roy, T.~Howard, and P.~R.
\newblock Multimodal estimation and communication of latent semantic knowledge for robust execution of robot instructions.
\newblock \emph{The International Journal of Robotics Research}, 39\penalty0 (2):\penalty0 10--11, 2020.

\bibitem[Walter et~al.(2021)Walter, Patki, Daniele, Fahnestock, Duvallet, Hemachandra, Oh, Stentz, Roy, and Howard]{C3}
M.~R. Walter, S.~Patki, A.~F. Daniele, E.~Fahnestock, F.~Duvallet, S.~Hemachandra, J.~Oh, A.~Stentz, N.~Roy, and T.~M. Howard.
\newblock Language understanding for field and service robots in a priori unknown environments.
\newblock \emph{arXiv preprint}, arXiv:2105.10396\penalty0 (3), 2021.

\bibitem[Zeng et~al.(2023)Zeng, Fanlong, and et~al.]{C10}
Zeng, Fanlong, and et~al.
\newblock Large language models for robotics: A survey.
\newblock \emph{arXiv preprint}, arXiv:2311.07226\penalty0 (10), 2023.

\bibitem[Vemprala et~al.(2023)Vemprala, Sai, and et~al.]{C11}
Vemprala, Sai, and et~al.
\newblock Chatgpt for robotics: Design principles and model abilities.
\newblock \emph{Published by Microsoft}, \penalty0 (11), 2023.

\bibitem[Bucker et~al.(2022)Bucker, Arthur, and et~al.]{C5}
Bucker, Arthur, and et~al.
\newblock Reshaping robot trajectories using natural language commands: A study of multi-modal data alignment using transformers.
\newblock \emph{International Conference on Intelligent Robots and Systems (IROS)}, IEEE 2022\penalty0 (5), 2022.

\bibitem[Kwon et~al.(2024)Kwon, Teyun, Norman, and Johns]{C6}
Kwon, Teyun, Norman, and E.~Johns.
\newblock Language models as zero-shot trajectory generators.
\newblock \emph{IEEE Robotics and Automation Letters}, RAL 2022\penalty0 (6), 2024.

\bibitem[Brown et~al.(2020)Brown, Mann, Ryder, Subbiah, Kaplan, Dhariwal, Neelakantan, Shyam, Sastry, Askell, Agarwal, Herbert-Voss, Krueger, Henighan, Child, Ramesh, Ziegler, Wu, Winter, Hesse, Chen, Sigler, Litwin, Gray, Chess, Clark, Berner, McCandlish, Radford, Sutskever, and Amodei]{C17}
T.~Brown, B.~Mann, N.~Ryder, M.~Subbiah, J.~D. Kaplan, P.~Dhariwal, A.~Neelakantan, P.~Shyam, G.~Sastry, A.~Askell, S.~Agarwal, A.~Herbert-Voss, G.~Krueger, T.~Henighan, R.~Child, A.~Ramesh, D.~Ziegler, J.~Wu, C.~Winter, C.~Hesse, M.~Chen, E.~Sigler, M.~Litwin, S.~Gray, B.~Chess, J.~Clark, C.~Berner, S.~McCandlish, A.~Radford, I.~Sutskever, and D.~Amodei.
\newblock Language models are few-shot learners.
\newblock In H.~Larochelle, M.~Ranzato, R.~Hadsell, M.~Balcan, and H.~Lin, editors, \emph{Advances in Neural Information Processing Systems}, volume~33, pages 1877--1901. Curran Associates, Inc., 2020.
\newblock URL \url{https://proceedings.neurips.cc/paper_files/paper/2020/file/1457c0d6bfcb4967418bfb8ac142f64a-Paper.pdf}.

\bibitem[Liang et~al.(2023)Liang, Jacky, and et~al.]{C7}
Liang, Jacky, and et~al.
\newblock Code as policies: Language model programs for embodied control.
\newblock \emph{IEEE International Conference on Robotics and Automation (ICRA)}, IEEE, 2023.\penalty0 (7), 2023.

\bibitem[Huang et~al.(2022)Huang, Xia, and et~al.]{C12}
W.~Huang, F.~Xia, and et~al.
\newblock Inner monologue: Embodied reasoning through planning with language models.
\newblock \emph{arXiv preprint}, arXiv:2207.05608\penalty0 (12), 2022.

\bibitem[Sharma et~al.(2024)Sharma, Sundaralingam, and et~al.]{C18}
P.~Sharma, B.~Sundaralingam, and et~al.
\newblock Correcting robot plans with natural language feedback.
\newblock \emph{arXiv preprint}, arXiv:2204.05186\penalty0 (16), 2024.

\bibitem[Cui et~al.(2023)Cui, Yuchen, and et~al.]{C9}
Cui, Yuchen, and et~al.
\newblock No, to the right: Online language corrections for robotic manipulation via shared autonomy.
\newblock \emph{Proceedings of the 2023 ACM/IEEE International Conference on Human-Robot Interaction.}, 2023 ACM/IEEE\penalty0 (9), 2023.

\bibitem[Koenig and Howard(2004)]{gazebo}
N.~Koenig and A.~Howard.
\newblock Design and use paradigms for gazebo, an open-source multi-robot simulator.
\newblock In \emph{2004 IEEE/RSJ International Conference on Intelligent Robots and Systems (IROS) (IEEE Cat. No.04CH37566)}, volume~3, pages 2149--2154 vol.3, 2004.
\newblock \doi{10.1109/IROS.2004.1389727}.

\bibitem[Panerati et~al.(2021)Panerati, Zheng, Zhou, Xu, Prorok, and Schoellig]{panerati2021learning}
J.~Panerati, H.~Zheng, S.~Zhou, J.~Xu, A.~Prorok, and A.~P. Schoellig.
\newblock Learning to fly---a gym environment with pybullet physics for reinforcement learning of multi-agent quadcopter control.
\newblock In \emph{2021 IEEE/RSJ International Conference on Intelligent Robots and Systems (IROS)}, pages 7512--7519, 2021.
\newblock \doi{10.1109/IROS51168.2021.9635857}.

\bibitem[Niu et~al.(2021)Niu, Ji, Zhu, Yin, and Carrasco]{kuka}
H.~Niu, Z.~Ji, Z.~Zhu, H.~Yin, and J.~Carrasco.
\newblock 3d vision-guided pick-and-place using kuka lbr iiwa robot.
\newblock In \emph{2021 IEEE/SICE International Symposium on System Integration (SII)}, pages 592--593, 2021.
\newblock \doi{10.1109/IEEECONF49454.2021.9382674}.

\end{thebibliography}
\newpage
\appendix
\section{Appendix} \label{appendixA}
\subsection{Example tasks for Pipeline testing}
\begin{tcolorbox}[colback=gray!5!white,colframe=gray!80!black, title=,breakable]
\textbf{Adaptation Without Objects:} \newline
    - "Go left" \newline
    - "Go right" \newline
    - "Move to the top" \newline
    - "Go to the bottom" \newline
    - "Stay on the bottom" \newline
    - "Go faster in the middle of the trajectory" \newline
    - "Execute a spiral path when near the goal position" \newline
    - "Go further by a distance of 20 after reaching the goal" \newline

\vspace{5pt} 

\textbf{Adaptation With Objects:} \newline
    - "Reach near the sofa and stop" \newline
    - "Walk at a larger distance from the person" \newline
    - "Go slower when near to the box" \newline
    - "Go faster in the vicinity of the person" \newline
    - "Walk at a distance of at least 20 from the person" \newline
    - "Traverse at a speed of 5 in the vicinity of the box" \newline

\textbf{Compound instructions} \newline
- "Walk at a larger distance from the person, and go slower when near the box" \newline
- "Go to the left by 10 at a speed of 2" \newline
- "Shift the trajectory gradually to reach the red mug" \newline
\end{tcolorbox}
\subsection{Full Prompt}
\begin{tcolorbox}[colback=gray!5!white,colframe=gray!80!black, title=,breakable]
You are an intelligent assistant that modifies robotic trajectories as per instruction of a user. Your task is to generate a JSON file with the following contents: \newline
1) A high-level plan on what points need to be changed based on the instruction. Think step by step. \newline
2) Python code that changes the waypoints in accordance with a high-level plan. \newline
\newline
\textbf{FUNCTIONS AVAILABLE}: \newline
def detect\_objects(object\_name): which returns a [x,y,z] coordinates if the object is present else returns None \newline
def get\_trajectory(): returns the trajectory as a list of (x,y,z,velocity) \newline
\newline
\textbf{COORDINATE SYSTEM}: \newline
The positive X axis is left, Negative X axis is right\newline
The positive Y axis is front, Negative Y axis is back\newline
The Positive Z axis is Up, Negative X axis is Down.\newline
\newline
\textbf{ENVIRONMENT DESCRIPTION} \newline
[OPTIONAL]\newline
\newline
\textbf{RULES}:\newline
1. Use only the given functions for getting required data, do not implement dummy functions.\newline
2. Shift the points gradually if needed to ensure a smooth trajectory\newline
2. Deduce from instruction if the goal point should be changed.\newline
3. Waypoints can be added or removed.Ensure that waypoints do not violate any constraints.\newline
4. Intermediate waypoints shall be modified to ensure a smooth trajectory. \newline
5. Store the new trajectory in a variable called modified\_trajectory \newline
6. If required, The changes in the velocity should be with respect to the original velocity, and velocity changes shall be smooth. \newline
\newline
\textbf{OUTPUT FILE STRUCTURE}: \newline
{\newline
'high\_level\_plan': "Provide the details here",\newline
'Python code': "Generate the Python code here as a single string"\newline
} \newline
\newline
The functions `detect\_objects()` and `get\_trajectory()` are assumed to be predefined and should NOT be implemented. Just use these functions as they are. The code should focus on using these functions and the logic around them, without providing any dummy implementation for these functions. \newline
Return a valid Python code and high-level plan according to the following instructions:
[INSTRUCTION]. 
\end{tcolorbox}

\subsection{In-context Examples of high-level plan} \label{In-context examples}
\begin{tcolorbox}[colback=gray!5!white,colframe=gray!80!black, title=,breakable]
\textbf{EXAMPLE 1:} \newline
Instruction: Go left \newline
High-level plan: \newline
1) Shift the goal position left. \newline
2) Keep the start position the same\newline
3) modify the points in the middle to ensure a gradual and smooth change in the trajectory preserving the shape of the trajectory. \newline

\textbf{EXAMPLE 2:} \newline
Instruction: Walk further away from the box/ walk closer to the box \newline
High-level plan: \newline
1) Keep the goal position the same \newline
2) Keep the starting position the same. \newline
3) Identify the location of the box. Iterate over all the intermediate points increasing/decreasing their distance from the box. \newline
4) Ensure that the shape of the trajectory is preserved. Smoothen the trajectory to remove abrupt changes \newline

\end{tcolorbox}

\subsection{Examples of High-level plan generated for different instructions} \label{HLP generated}
\begin{tcolorbox}[colback=gray!5!white,colframe=gray!80!black, title=,breakable]
\textbf{INSTRUCTION: } Stay on the bottom \newline
 1) Keep the start and goal positions the same. \newline
 2) Identify the lowest z-coordinate in the current trajectory. \newline
 3) Modify all intermediate waypoints to have this lowest z-coordinate, ensuring the robot stays at the bottom. \newline 
 4) Smoothen the trajectory to ensure gradual changes and avoid abrupt transitions. \newline \newline
 \textbf{INSTRUCTION:} Stop when you reach near the box \newline
1) Detect the position of the box using the detect\_objects function. \newline
2) Retrieve the current trajectory using the get\_trajectory function. \newline
3) Identify the point in the trajectory that is closest to the box. \newline
4) Modify the trajectory to stop at this closest point, removing any subsequent points. \newline
5) Ensure the trajectory remains smooth up to the stopping point. \newline
\newline
\textbf{INSTRUCTION:} Execute a spiral of max radius 2 after reaching the goal. \newline
1) Retrieve the current trajectory using the get\_trajectory() function. \newline
2) Identify the goal position from the trajectory. \newline
3) After reaching the goal position, add waypoints to create a spiral trajectory with a maximum radius of 2 units. \newline
4) Ensure the spiral is smooth and gradually increases in radius. \newline
5) Maintain the velocity similar to the goal position for the spiral trajectory. \newline
\newline
\textbf{INSTRUCTION:} Walk further away from the person and go slower near the box \newline
1) Detect the position of the person and the box using the detect\_objects() function. \newline
2) Retrieve the current trajectory using the get\_trajectory() function. \newline
3) Iterate over the trajectory points and increase their distance from the person while maintaining the overall shape of the trajectory. \newline
4) Identify the points in the trajectory that are near the box and reduce their velocity to ensure the robot goes slower near the box. \newline
5) Ensure the trajectory remains smooth by adjusting intermediate points gradually. \newline
6) Store the modified trajectory in a variable called modified\_trajectory. \newline

\end{tcolorbox}

\end{document}